%% file: main.tex
\pdfoutput=1
\documentclass[10pt,twocolumn,letterpaper]{article}

\usepackage[pagenumbers]{cvpr} 

\usepackage{graphicx}
\usepackage{amsmath}
\usepackage{amssymb}
\usepackage{booktabs}
\usepackage{graphicx}
\usepackage{multirow}
\usepackage[export]{adjustbox}
\usepackage[accsupp]{axessibility}  

\usepackage{pifont}
\usepackage{xcolor}
\definecolor{deemph}{gray}{0.6}

\newcommand{\cmark}{\ding{51}\xspace}%
\newcommand{\cmarkg}{\textcolor{lightgray}{\ding{51}}\xspace}%
\newcommand{\xmark}{\ding{55}\xspace}%
\newcommand{\xmarkg}{\textcolor{lightgray}{\ding{55}}\xspace}%

\usepackage[pagebackref,breaklinks,colorlinks]{hyperref}

\usepackage[capitalize]{cleveref}
\crefname{section}{Sec.}{Secs.}
\Crefname{section}{Section}{Sections}
\Crefname{table}{Table}{Tables}
\crefname{table}{Tab.}{Tabs.}

\def\cvprPaperID{130} 
\def\confName{CVPR}
\def\confYear{2022}

\begin{document}

\title{NAFSSR: Stereo Image Super-Resolution Using NAFNet}

\author{ Xiaojie Chu$^{2}$\thanks{Equal contribution.}\hspace{20pt}  Liangyu Chen$^1$\footnotemark[1]  \hspace{20pt} Wenqing Yu$^1$ \\ 
{$^1 $} MEGVII Technology \hspace{20pt} {$^2 $} Peking University \\
{\tt\small chuxiaojie@stu.pku.edu.cn \hspace{20pt} \{chenliangyu,yuwenqing\}@megvii.com}
}
\maketitle

\maketitle

\input{chaps/abstract}
\input{chaps/introduction}
\input{chaps/related_works}

\input{chaps/methods}

\input{chaps/experiments}

\input{chaps/challenge}
\input{chaps/conclusion}
{\small
\bibliographystyle{ieee_fullname}
\bibliography{main}
}

\end{document}

%% file: chaps/abstract.tex
\begin{abstract}
Stereo image super-resolution aims at enhancing the quality of super-resolution results by utilizing the complementary information provided by binocular systems. To obtain reasonable performance, most methods focus on finely designing modules, loss functions, and \etc to exploit information from another viewpoint. This has the side effect of increasing system complexity, making it difficult for researchers to evaluate new ideas and compare methods. This paper inherits a strong and simple image restoration model, NAFNet, for single-view feature extraction and extends it by adding cross attention modules to fuse features between views to adapt to binocular scenarios. The proposed baseline for stereo image super-resolution is noted as NAFSSR. Furthermore, training/testing strategies are proposed to fully exploit the performance of NAFSSR. Extensive experiments demonstrate the effectiveness of our method. In particular, NAFSSR outperforms the state-of-the-art methods on the KITTI 2012, KITTI 2015, Middlebury, and Flickr1024 datasets. With NAFSSR, we won 1st place in the NTIRE 2022 Stereo Image Super-resolution Challenge. Codes and models will be released at \url{https://github.com/megvii-research/NAFNet}.
\end{abstract}

%% file: chaps/introduction.tex
\section{Introduction}
Stereo image super-resolution (SR), which aims at reconstructing high-resolution (HR) details from a pair of low-resolution (LR) left and right images, has attracted much attention in recent years. To solve this task, both context information within a single view (\ie intra-view information) and information between left and right image (\ie cross-view information) are crucial~\cite{ying2020stereo}. 
 On the one hand, recent works in stereo image SR~\cite{wang2021symmetric, zhu2021cross, dai2021feedback} mainly focus on the finely designing novel network architectures, losses, and \etc to effectively incorporate additional information from another viewpoint, as the cross-view information provided by binocular systems enhances the image quality. But the system complexity is increasing, which may hinder the convenient analysis and comparison of methods. On the other hand, remarkable progress in single image restoration has been witnessed with deep learning techniques, \eg Transformer-based SwinIR~\cite{liang2021swinir} outperforms state-of-the-art methods on single-image SR. NAFNet~\cite{chen2022simple} achieves state-of-the-art performance without nonlinear activation functions on denoising and deblurring tasks. However, these single image restorers are suboptimal for stereo image SR as they cannot utilize the cross-view information.

\begin{figure}[t]
	\centering
	\includegraphics[width=0.47\textwidth]{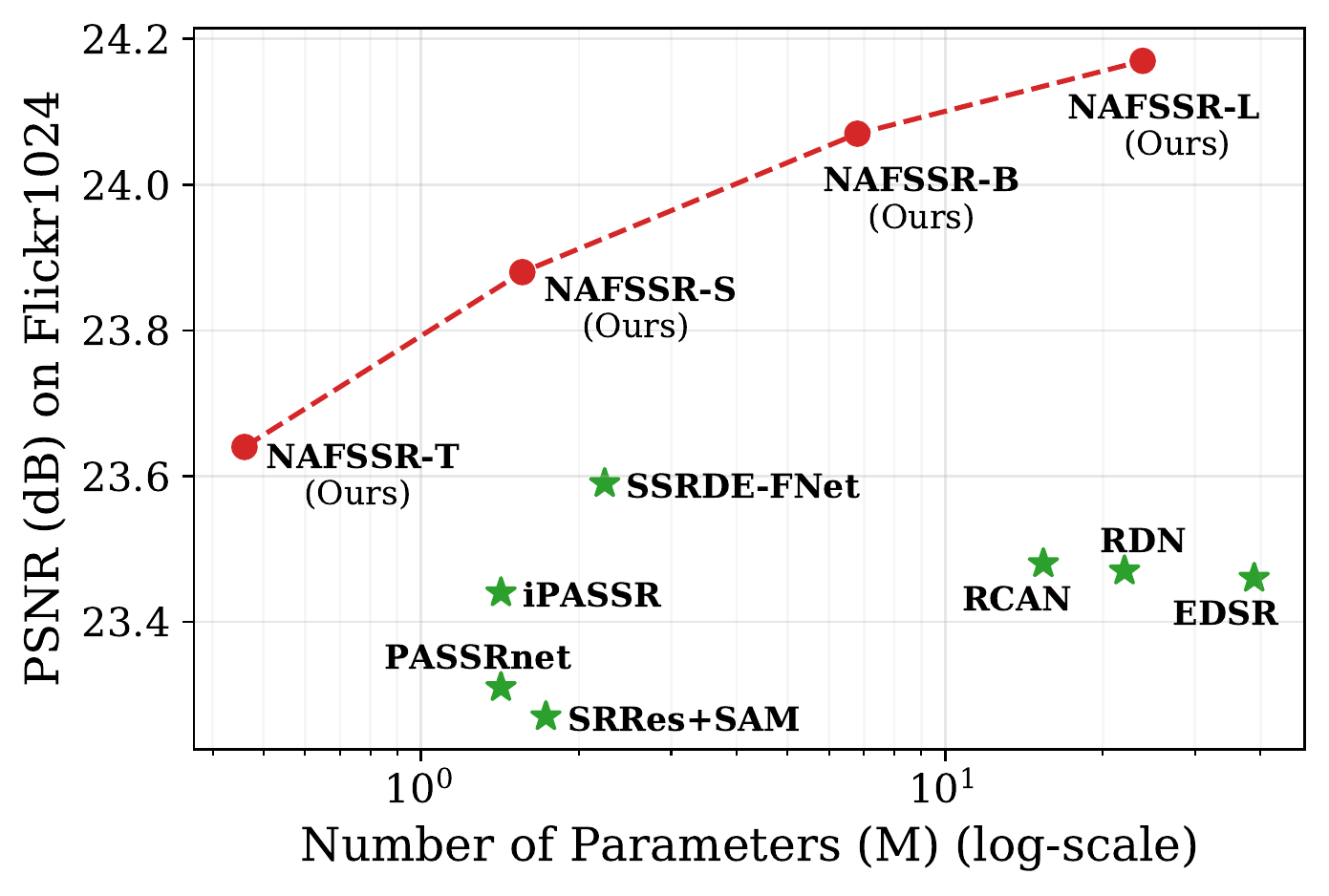}
	\vspace{-4mm}
	\caption{Parameters vs. PSNR of models for 4$\times$ stereo SR on Flickr1024~\cite{wang2019flickr1024} test set.
	Our NAFSSR families achieve the state-of-the-art performance with up to 79\% of parameter reduction.
   }
	\label{fig:nafssr-params}
\end{figure}

Inspired by NAFNet~\cite{chen2022simple} which achieves competitive performance on single image restoration tasks with low system complexity, we propose a novel baseline for stereo image SR, NAFSSR, by adding simple cross attention modules to NAFNet.
It can fully utilize both intra-view information and cross-view information to achieve the competitive performance of stereo super-resolution. Specifically, we stack NAFNet blocks (NAFBlocks for short) and extract intra-view features for both views in a weight-sharing manner.
It inherits the strong representation (within the viewpoint) of NAFNet. 
Specifically, to further improve the representation of NAFNet, we propose stereo cross-attention module (SCAM) to attend and fuse the left/right viewpoint features. It first computes bidirectional cross attention from left to right and right to left views, and then fuses the interacted cross-view features with intra-view features.
In contrast to the original cross-attention used in a standard Transformer decoder~\cite{vaswani2017attention}, which attends to all locations in an image, our stereo cross-attention attends to corresponding features along the horizontal epipolar line, following~\cite{wang2019learning,wang2021symmetric}. 

Although NAFSSR has strong representational power, it may suffer from overfitting due to the lack of data for the stereo SR task. To solve this, we adopt stochastic depth\cite{huang2016deep} as regularization and channel shuffle (\ie, shuffle the RGB channels of input images randomly) as data augmentation during the training phase. Besides, we reveal that there is also the train/test inconsistency issue mentioned in TLSC~\cite{chu2021tlsc} in the stereo SR task. Thus we adopt TLSC~\cite{chu2021tlsc} in the testing phase to alleviate the inconsistency issue. These training/testing strategies, together with NAFSSR, constitute a baseline for the stereo SR task. As shown in Figure~\ref{fig:nafssr-params}, our NAFSSR families have better performance and parameters trade-off than existing methods.

Our contributions can be summarized as follows:
\begin{itemize}
\vspace{-2mm}
	\item  We analyze the drawbacks of existing methods and propose NAFSSR, which is simple and easily implemented. It inherits the advantages of NAFNet's simplicity and power, and uses the characteristics of the stereo SR task to improve the representation through a simple stereo cross-attention module.
\vspace{-2mm}
	\item  Based on NAFSSR, we design its training/testing strategies, thus addressing the obstacles to its competitive performance on the stereo SR task. The strategies together with NAFSSR constitute a strong baseline for this task: the baseline achieves the state-of-the-art performance with fewer parameters (Figure~\ref{fig:nafssr-params}) and faster inference speed (Table~\ref{tab:runtime}).
\vspace{-2mm}
    \item Extensive experiments are conducted to demonstrate the effectiveness of our proposed NAFSSR. With the help of NAFSSR, we won 1st place in the NTIRE 2022 Stereo Image Super-resolution Challenge~\cite{Wang2022NTIRE}.
\end{itemize}

%% file: chaps/related_works.tex
\section{Related Works}

\subsection{Single Image Super-resolution}
Single image restoration tasks, \eg, image super-resolution (SR), aim at reconstructing high-quality images by using only intra-view information from low-quality input.
Deep learning-based methods have dominated single image super-resolution tasks since the pioneering work of Super-Resolution Convolutional Neural Network (SRCNN~\cite{dong2014learning}).
More complicated neural network architecture designs have been presented to improve model representation ability by increasing the depth and width of models~\cite{kim2016accurate}, applying residual~\cite{lim2017enhanced, zhang2018image} and dense~\cite{zhang2018residual} connections, as well as introducing different attention mechanism (\eg, channel attention~\cite{zhang2018image, dai2019second, magid2021dynamic}, channel-spatial attention~\cite{dai2019image, niu2020single, mei2021image, liang2021swinir}). 
Specifically, SwinIR~\cite{liang2021swinir} proposes a Swin Transformer-based image restoration method and achieves state-of-the-art performance on single image SR.
In this paper, we extend NAFNet~\cite{chen2022simple}, a simple baseline with competitive performance on single image restoration tasks, to stereo image SR task.

\subsection{Stereo Super-Resolution}
Stereo super-resolution task aims at reconstructing high-resolution details of a pair of low-resolution images on the left and right views.
StereoSR~\cite{jeon2018enhancing} learns a mapping between continuous parallax shifts and a high-resolution image by jointly training two cascaded sub-networks for luminance and chrominance, respectively. 
To handle different stereo images with large disparity variations, PASSRnet~\cite{wang2019learning} introduces a parallax-attention mechanism with a global receptive field along the epipolar line.
Ying~\etal~\cite{ying2020stereo} propose a stereo attention module (SAM) to extend pre-trained single image SR networks for stereo image SR. 
StereoIRN~\cite{yan2020disparity} introduces two disparity attention losses and uses a pre-trained disparity flow network to align two views features.
Song~\etal~\cite{song2020stereoscopic} propose self and parallax attention mechanism for simultaneously aggregating information from its own image and the counterpart stereo image.
To effectively interact cross-view information, iPASSR~\cite{wang2021symmetric} propose symmetric bi-directional parallax attention module (biPAM) and an inline occlusion handling scheme to exploit symmetry cues for stereo image SR.
CVCnet~\cite{zhu2021cross} integrates cross view spatial features from both global and local perspectives.
SSRDE-FNet~\cite{dai2021feedback} simultaneously handles the stereo image SR and disparity estimation in a unified framework and interacts two tasks in a mutually boosted way. 

We also design a simple stereo cross-attention module to extend single image restoration networks for stereo image SR.
In contrast to SAM~\cite{ying2020stereo}, which uses single image SR models pretrained on extra datasets and only fine-tunes on stereo datasets with multiple losses, our NAFSSR is trained directly on stereo images from scratch with only L1 loss.

\begin{figure*}[t]
\centering
\includegraphics[width=0.96  \textwidth]{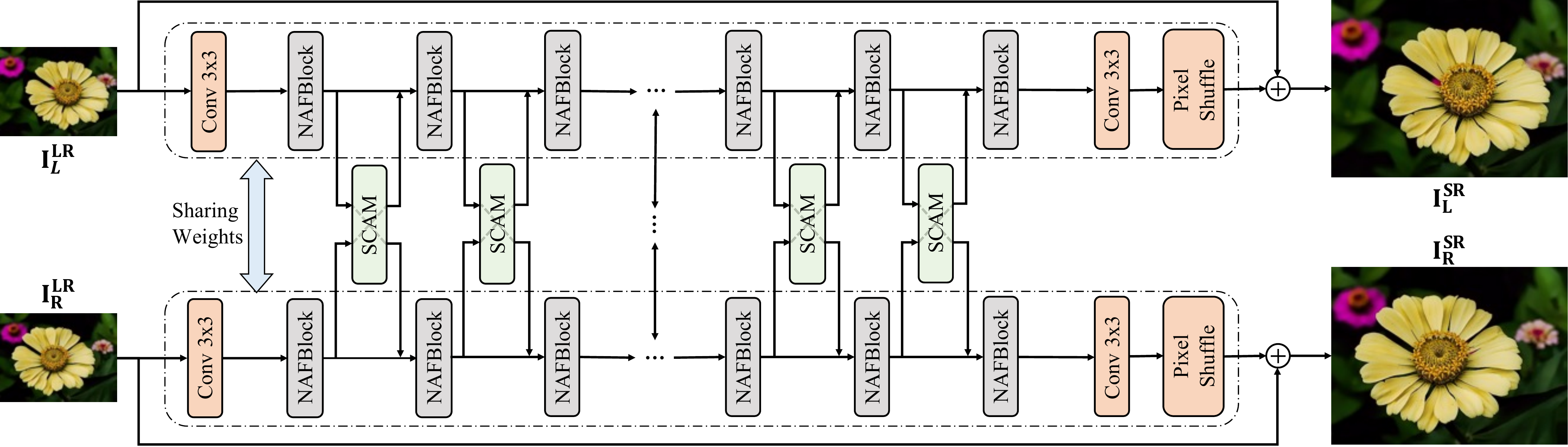}
\vspace{-2mm}
\caption{The overall architecture of NAFSSR.
SCAM represents Stereo Cross Attention Module (shown in Figure~\ref{fig.attention}).
}
\vspace{-2mm}
\label{fig.overall}
\end{figure*}

\begin{figure}[t]
\centering
\includegraphics[width=0.47\textwidth]{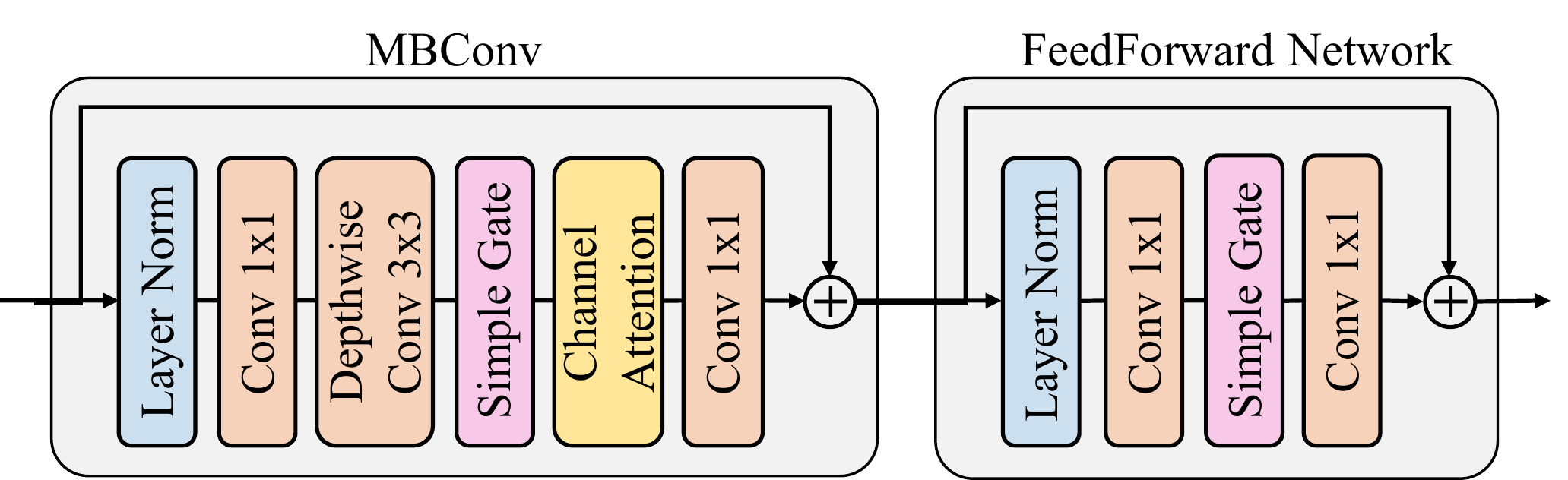}
\vspace{-2mm}
\caption{NAFBlock. Simple Gate and Channel Attention Module are shown in Equation~\ref{eq:sg} and Equation~\ref{eq:ca}, respectively. }
\label{fig.block}
\vspace{-2mm}
\end{figure}

\subsection{Training and Testing Strategies}
Regularizations (\eg, weight decay~\cite{wightman2021resnet}, dropout and stochastic depth~\cite{huang2016deep}) are widely used to improve model performance in high-level computer vision tasks~\cite{wightman2021resnet}.
However, there is still no consensus on whether regularization techniques should be used in image super-resolution (SR) tasks. For example, Lin~\etal~\cite{lin2022revisiting} discover that underfitting is still the main issue limiting the model capability of RCAN~\cite{zhang2018image}. On the contrary, Kong~\etal~\cite{kong2021reflash} demonstrate that proper use of dropout~\cite{hinton2012improving} benefits SR networks by preventing overfitting to a specific degradation.
In this paper, we find that the proposed networks (except the smallest one) are overfitting to the stereo training data, so we use stochastic depth to improve their generality.

%% file: chaps/methods.tex
\section{Method}
In this section, we introduce our method in details.
We first describe the architecture of our network in Section~\ref{sec:arch}, then discuss the training and testing strategies throughout the paper in Section~\ref{sec:training} and ~\ref{sec:testing}, respectively.
\subsection{Network Architecture}\label{sec:arch}
\subsubsection{Overall Framework}
An overview of our proposed NAFNet-based~\cite{chen2022simple} Stereo Super-Resolution network (NAFSSR) is illustrated in Figure~\ref{fig.overall}. NAFSSR takes the low-resolution stereo image pair as input and super-resolves both left and right view images.
Two weight-sharing networks (stacked by NAFBlock) extract the intra-view features of the left and the right images separately. And Stereo Cross-Attention Modules (SCAMs) are provided to fuse features extracted from the left and the right images. 
In detail, NAFSSR can be divided into three parts: intra-view feature extraction, cross-view feature fusion, and reconstruction.

\textbf{Intra-view feature extraction and reconstruction.}
In the beginning, a $3\times3$ convolution layer is used to map the input image space to a higher dimensional feature space. Then, $N$ NAFBlocks are used for deep intra-view feature extraction. The details of NAFBlock are described in Section~\ref{sec:NAFBlock}.
After feature extraction, a $3\times3$ convolution layer followed by a pixel shuffle layer~\cite{shi2016real} is used to upsample the feature by a scale factor of $s$. 
Furthermore, to alleviate the burden of feature learning, we use global residual learning and predict only the residual between the bilinearly upsampled low-resolution image and the ground-truth high-resolution image~\cite{liang2022vrt}. 

\textbf{Cross-view feature fusion.}
To interact with cross-view information, we insert SCAM after each NAFBlock.
It uses stereo features generated by previous NAFBlocks as inputs to perform bidirectional cross-view interactions, and outputs interacted features fused with input intra-view features. 
The details of SCAM are described in Section~\ref{sec:fusion}.

\subsubsection{NAFBlock}\label{sec:NAFBlock}
The NAFBlock is introduced by NAFNet~\cite{chen2022simple}, and its details are shown in Figure~\ref{fig.block}. It should be noticed that there are no nonlinear activation functions in it. 
NAFBlock consists of two parts: (1) Mobile convolution module (MBConv) based on point-wise and depth-wise convolution with channel attention (simplified SE~\cite{hu2018squeeze}); (2) a feed-forward network (FFN) module that has two fully-connected layers (implemented by point-wise convolution). The LayerNorm (LN~\cite{ba2016layer}) layer is added before both MBConv and FFN, and the residual connection is employed for both modules. The whole process is formulated as:
\begin{equation}
\begin{aligned}
&\mathbf{X}=\operatorname{MBConv}(\operatorname{LN}(\mathbf{X}))+\mathbf{X} \\
&\mathbf{X}=\operatorname{FFN}(\operatorname{LN}(\mathbf{X}))+\mathbf{X}
\end{aligned}
\end{equation}

The main differences between NAFBlock and original blocks (\eg, MBConv in MobileNetV3~\cite{howard2019searching} and FFN in Transformer~\cite{vaswani2017attention}) lie in the simple gate mechanism, which makes block nonlinear activation free. Specifically, NAFBlock uses SimpleGate unit to replace nonlinear activation (\eg, ReLU, GELU). 
Given an input $\mathbf{X} \in \mathbb{R}^{H \times W \times C}$, SimpleGate first split the input into two features $\mathbf{X_1}, \mathbf{X_2} \in \mathbb{R}^{H \times W \times C/2}$ along channel dimension. Then, it computes the output with linear gate as:
\begin{equation}
\operatorname{SimpleGate}(\mathbf{X})=\mathbf{X_1} \odot \mathbf{X_2},\label{eq:sg}
\end{equation}
where $\odot$ represents element-wise multiplication. The SimpleGate unit is added after depth-wise convolution and between two fully-connected layers.

\subsubsection{Stereo Cross Attention Module}\label{sec:fusion}
\begin{figure}[t]
\centering
\includegraphics[width=0.4 \textwidth]{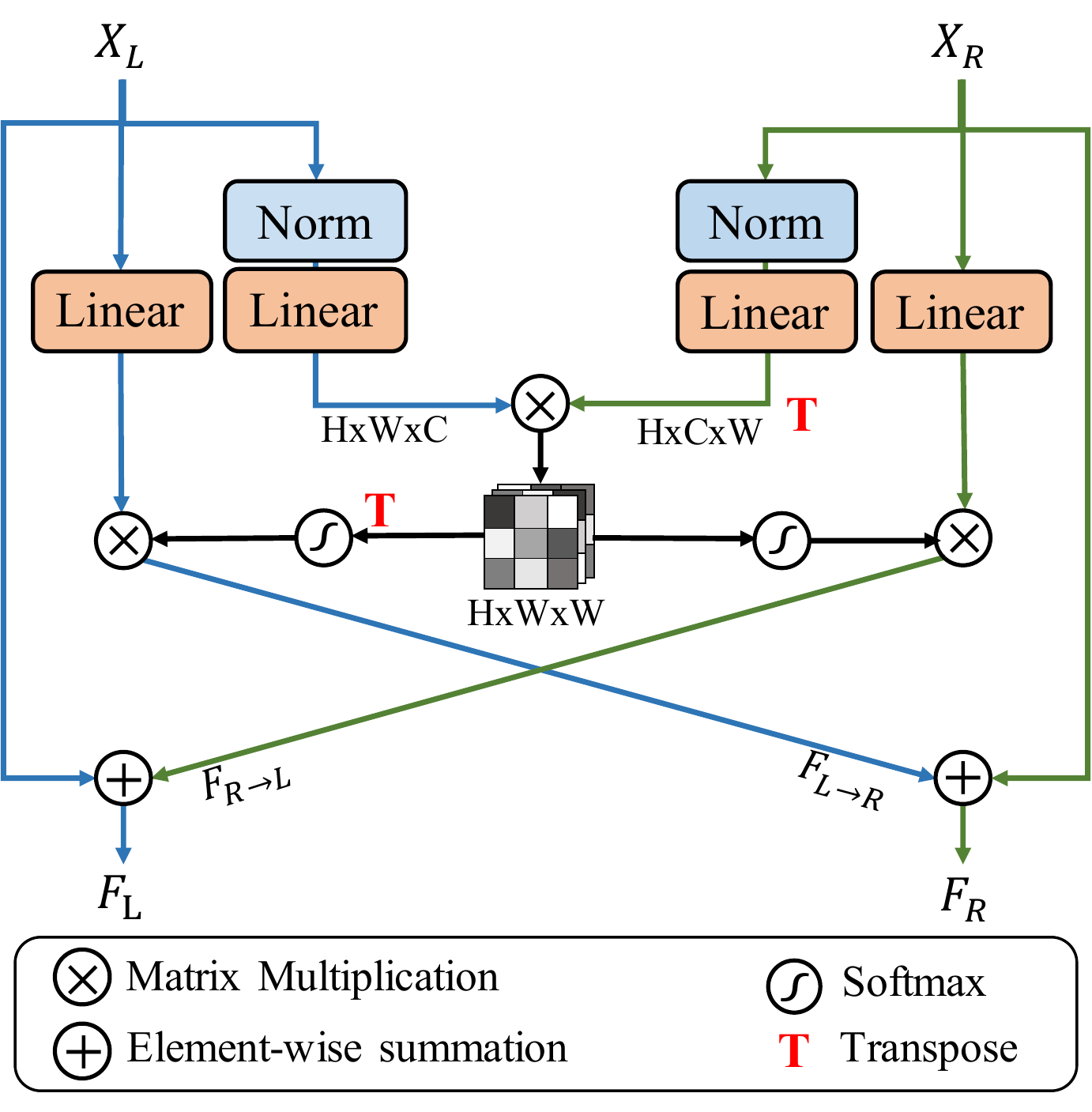}
\vspace{-2mm}
\caption{Stereo Cross Attention Module (SCAM). It fuses the features of the left and right views.
}
\vspace{-2mm}
\label{fig.attention}
\end{figure}

The details of the proposed Stereo Cross Attention Module (SCAM) are shown in Figure~\ref{fig.attention}.
It is based on Scaled Dot-Product Attention~\cite{vaswani2017attention}, which computes the dot products of the query with all keys and applies a softmax function to obtain the weights on the values:
\begin{equation}
\operatorname{Attention}(\mathbf{Q}, \mathbf{K}, \mathbf{V})=\operatorname{softmax}\left(\mathbf{Q} \mathbf{K}^{T}/\sqrt{C}\right) \mathbf{V}
\end{equation}
where $\mathbf{Q} \in \mathbb{R}^{H \times W \times C}$ is \textit{query} matrix projected by source intra-view feature (\eg, left-view), and $\mathbf{K}, \mathbf{V} \in \mathbb{R}^{H \times W \times C}$ are \textit{key}, \textit{value} matrices projected by target intra-view feature (\eg, right-view). Here, $H, W, C$ represent height, width and number of channels of feature map.
Since stereo images are highly symmetric under epipolar constraint~\cite{wang2021symmetric}, we use the same $\mathbf{Q}$ and $\mathbf{K}$ to represent each intra-view features, and calculates the correlation of cross-view features on a horizontal line (\ie, along $W$ dimension).
In detail, given the input stereo intra-view features $\mathbf{X_{L}}, \mathbf{X_{R}} \in \mathbb{R}^{H \times W \times C}$, we can get layer normalized stereo features $\mathbf{\Bar{X}_L} = \operatorname{LN}(\mathbf{X_L})$ and $\mathbf{\Bar{X}_R} = \operatorname{LN}(\mathbf{X_R})$. 
Then, we calculate bidirectional cross-attention between left-right views by:
\begin{equation} 
\begin{split}
\mathbf{F_{R\rightarrow L}} &= \operatorname{Attention}(\mathbf{W^L_{1}}\mathbf{\Bar{X}_L}, \mathbf{W^R_{1}}\mathbf{\Bar{X}_R}, \mathbf{W^R_{2}}\mathbf{X_R}), \\
\mathbf{F_{L\rightarrow R}} &= \operatorname{Attention}(\mathbf{W^R_{1}}\mathbf{\Bar{X}_R}, \mathbf{W^L_{1}}\mathbf{\Bar{X}_L}, \mathbf{W^L_{2}}\mathbf{X_L}),
\end{split}
\end{equation}
where $\mathbf{W^L_{1}},\mathbf{W^R_{1}}, \mathbf{W^L_{2}}$ and $\mathbf{W^R_{2}}$ are projection matrices.
Note that we can calculate the left-right attention matrix only once to generate both $\mathbf{F_{R\rightarrow L}}$ and $\mathbf{F_{L\rightarrow R}}$ (as shown in Figure~\ref{fig.attention}).
Finally, the interacted cross-view information $\mathbf{F_{R\rightarrow L}}, \mathbf{F_{L\rightarrow R}}$ and intra-view information $\mathbf{X_L}, \mathbf{X_R}$ are fused by element-wise addition:
\begin{equation} 
\begin{split}
\mathbf{F_L} &= \gamma_L\mathbf{F_{R\rightarrow L}} + \mathbf{X_L}, \\
\mathbf{F_R} &= \gamma_R\mathbf{F_{L\rightarrow R}} + \mathbf{X_R},
\end{split}
\end{equation}
where $\gamma_L$ and $\gamma_R$ are trainable channel-wise scale and initialized with zeros for stabilizing training.

\subsection{Training Strategies}\label{sec:training}
\paragraph{Combat overfitting.} 
In stereo image SR tasks, it is common practice to train models with small patches cropped from full-resolution images~\cite{wang2019learning, wang2021symmetric, dai2021feedback}. These patches are randomly flipped horizontally and vertically for data augmentation. To further utilize the training data, we introduce \textbf{Channel Shuffle}: which randomly shuffles the RGB channels of input images for color augmentation. In addition, we adopt stochastic depth\cite{huang2016deep} as regularization.

\paragraph{Loss.}
For simplicity, we only use the pixel-wise L1 distance between the super-resolution and ground-truth stereo images:
\begin{equation}
\mathcal{L}=\left\|\mathbf{I}_{\text{L}}^{\text{SR}}-\mathbf{I}_{\text{L}}^{\text{HR}}\right\|_{1}+\left\|\mathbf{I}_{\text{R}}^{\text{SR}}-\mathbf{I}_{\text{R}}^{\text{HR}}\right\|_{1}
\end{equation}
where $\mathbf{I}_{\text{L}}^{\text{SR}}$ and $\mathbf{I}_{\text{R}}^{\text{SR}}$ represent the super-resolution left and right images generated by model respectively, and $\mathbf{I}_{\text{L}}^{\text{HR}}$ and $\mathbf{I}_{\text{R}}^{\text{HR}}$ represent their ground-truth high-resolution images. 

\subsection{Train-test Inconsistency}\label{sec:testing}
Chu~\etal~\cite{chu2021tlsc} discover that the distribution of image-based features during inference differs from that of patch-based features during training, and show that this train-test inconsistency harms model performance on debluring, denosising, deraining, and dehazing tasks.
For stereo image super-resolution task, the regional range of the inputs for training and inference also varies greatly, \eg, the range of region for each patch is only 4.5\% of low-resolution images ($30 \times 90$ vs. $300 \times 200$) in Flickr1024 dataset.  
This prompts us to check the potential train-test inconsistency issue 
of channel attention used in our network. 

In detail, given input features $X$, the channel attention ($\operatorname{CA}$) first aggregates global spatial information using global average pooling ($\operatorname{pool}$), and then redistributes the pooled information to input features as follows:
\begin{equation}
\operatorname{CA}(\mathbf{X})=\mathbf{X} * \mathbf{W} \operatorname{pool}(\mathbf{X}), \label{eq:ca}
\end{equation}
where $\mathbf{W}$ represents learnable matrix and $*$ is a channel-wise product operation. 
We apply TLSC~\cite{chu2021tlsc} to $\operatorname{CA}$ in Equation~\ref{eq:ca}, which converts $\operatorname{pool}$ operation from global average pooling to local average pooling during inference, allowing it to extract representations based on local spatial region of features as in training phase. According to~\cite{chu2021tlsc}, the local size for pooling is simply set to 1.5$\times$ the size of the training patch.

\vspace{-1mm}

%% file: chaps/experiments.tex
\section{Experiments}
\subsection{Implementation Details}\label{sec:NTIRE.details}
\textbf{Evaluation Metrics.}
Peak signal-to-noise ratio (PSNR) and structural similarity (SSIM) were used as quantitative metrics. 
These metrics are calculated on RGB color space with a pair of stereo images (\ie, $\left(\textit{Left}+\textit{Right}\right)/2$).

\textbf{Architecture.}
As shown in Table~\ref{tab:arch}, we construct 4 different size of NAFSSR networks by adjusting the number of channels and blocks, which are named NAFSSR-T (Tiny), NAFSSR-S (Small), NAFSSR-B (Base) and NAFSSR-L (Large).  Besides, we use TLSC~\cite{chu2021tlsc} during inference as described in Section~\ref{sec:testing}.
\begin{table}[]
\caption{Architecture Variants of NAFSSR.}
\vspace{-3mm}
\label{tab:arch}
\centering
\begin{tabular}{cccc}
\toprule
Models & \#Channels & \#Blocks & \#Params \\ \midrule
NAFSSR-T & $C=48$  & $N=16$ & 0.46M\\
NAFSSR-S & $C=64$ & $N=32$ & 1.56M \\
NAFSSR-B & $C=96$ & $N=64$ & 6.80M \\
NAFSSR-L & $C=128$ & $N=128$ & 23.83M\\
\bottomrule
\end{tabular}
\vspace{-2mm}
\end{table}

\textbf{Training.} All models are optimized by the AdamW with $\beta_1$ = 0.9 and $\beta_2$ = 0.9 with weight decay 0 by default.
The learning rate is set to $3 \times 10^{-3}$, and decreased to $1 \times 10^{-7}$ with cosine annealing strategy~\cite{loshchilov2016sgdr}.
If not specified, models are trained on $40 \times 100$ patches with a batch size of 32 for $1\times 10^5$ iterations.
We apply skip-init\cite{de2020batch} in our network, which may facilitate the training process.
Data augmentation is implemented as described in Section~\ref{sec:training}. To overcome the overfitting issue, we use stochastic depth~\cite{huang2016deep} with 0.1, 0.2 and 0.3 probability for NAFSSR-S, NAFSSR-B and NAFSSR-L, respectively. In particular, since our lightweight model NAFSSR-T encounters underfitting rather than overfitting, it uses $4\times$ training iterations without stochastic depth.

\textbf{Datasets.}
We use the training dataset and validation dataset provided by NTIRE Stereo Image Super-Resolution Challenge~\cite{Wang2022NTIRE}.
In detail, we use 800 stereo images from the training set of Flickr1024~\cite{wang2019flickr1024} dataset as the training data and 112 stereo images in the validation set of the Flickr1024~\cite{wang2019flickr1024} dataset as the validation set. The low-resolution images are generated by bicubic downsampling.

\subsection{Ablation Study}

\begin{table}[]
\centering
\setlength{\tabcolsep}{4.5pt}
\caption{4$\times$ SR results (PSNR) achieved on the Flickr1024~\cite{wang2019flickr1024} dataset by NAFSSR-S with different number of SCAMs.}
\vspace{-3mm}
\label{tab:scam}
\begin{tabular}{ccccccc}
\toprule
\#SCAM     & 0     & 1     & 4     & 8     & 16    & 32    \\
\midrule
PSNR         & 23.56 & 23.74 & 23.76 & 23.79 & 23.82 & 23.85 \\
$\Delta$PSNR & -     & +0.18 & +0.20 & +0.23 & +0.26 & +0.29 \\
\bottomrule
\end{tabular}
\vspace{-2mm}
\end{table}

\paragraph{Stereo Cross-Attention Module.} Here, we take NAFSSR-S without Stereo Cross-Attention Module (SCAM) as a naive baseline to investigate the impact of the proposed SCAM on the model performance. In this experiment, we apply different number of SCAM to the naive baseline, ranging from 0 to 32. In detail, we use SCAM after a specific number of NAFBlocks in the middle of the naive baseline. 
Note that our naive baseline (with 0 SCAM) only uses single-view information. In contrast, our NAFSSR-S (with 32 SCAMs) interacts with cross-view information after every NAFBlocks.

As demonstrated by the results in Table~\ref{tab:scam}, our SCAM offers significant performance improvements compared to the baseline.
The more number of SCAMs, the better performance. Compared to the naive baseline that uses only intra-view information, the PSNR on the Flickr1024 dataset can be improved by 0.18 dB with only one SCAM and by 0.29 dB with 32 SCAMs. 
These results indicate the importance of incorporating both cross-view information (introduced by our SCAM) and intra-view information (extracted by the NAFBlock).

\begin{table}[t]
\centering
\caption{4$\times$ SR results (PSNR) achieved on Flickr1024~\cite{wang2019flickr1024} by NAFSSR-S trained with different data augmentations. hflip and vflip represent horizontal flip and vertical flip, respectively.}
\label{tab:aug}
\vspace{-3mm}
\begin{tabular}{ccccc}
\toprule
hflip & vflip & channel shuffle & PSNR  & $\Delta$PSNR \\
\midrule
\xmarkg     &  \xmarkg    &      \xmarkg         & 23.43 & -            \\
\midrule
\cmark     &    \xmarkg   &   \xmarkg              & 23.64 & +0.21        \\
 \xmarkg     & \cmark     &       \xmarkg          & 23.63 & +0.20        \\
  \xmarkg    &   \xmarkg    & \cmark               & 23.62 & +0.19        \\ \midrule
\cmark     & \cmark     &    \xmarkg             & 23.73 & +0.30        \\
\cmark     & \cmark     & \cmark               & 23.82 & +0.39        \\
\bottomrule
\end{tabular}
\vspace{-2mm}
\end{table}

\begin{table*}[t]
    \centering
    \small
\caption{Effect of stochastic depth~\cite{huang2016deep} and TLSC~\cite{chu2021tlsc} to PSNR values of different models for $4\times$ SR on different datasets.}
\label{tab:droppath}
\vspace{-3mm}
 
    \begin{tabular}{ccclllll}
\toprule 
& Training & Test & \multicolumn{1}{c}{In-distribution}  & \multicolumn{4}{c}{Out-distribution} \\ 
\cmidrule(lr){2-2} \cmidrule(lr){3-3} \cmidrule(lr){4-4} \cmidrule(lr){5-8} 
Model  &Stoch. Depth & TLSC& \multicolumn{1}{c}{Flickr1024~\cite{wang2019learning}} & \multicolumn{1}{c}{KITTI 2012~\cite{geiger2012we}} & \multicolumn{1}{c}{KITTI 2015~\cite{menze2015object}} & \multicolumn{1}{c}{Middlebury~\cite{scharstein2014high}} & Average \\ \midrule
\multirow{3}{*}{NAFSSR-S}  & \cmarkg & \cmarkg& 23.85     & 26.91      & 26.74      & 29.63  & 27.76     \\
 & \xmark &\cmarkg & 23.82 ($-$0.03)      & 26.88 ($-$0.03)     & 26.71 ($-$0.03)     & 29.61 ($-$0.02)  & 27.73 ($-$0.03)    \\
 &  \cmarkg& \xmark& 23.78 ($-$0.07)      & 26.86 ($-$0.05)      & 26.67 ($-$0.07)     & 29.54  ($-$0.09) & 27.69 ($-$0.07)    \\ 
\midrule
\multirow{3}{*}{NAFSSR-B}  &\cmarkg & \cmarkg& 24.10      & 27.05 &
26.89  & 29.93  &  27.96    \\
 &\xmark & \cmarkg& 
23.98 ($-$0.11) &
26.92 ($-$0.13) &
26.70 ($-$0.19) &
29.78 ($-$0.15)& 
27.80 ($-$0.16)\\
&\cmarkg &\xmark & 
24.01 ($-$0.09)&
27.00 ($-$0.05)&
26.80 ($-$0.09)&
29.81 ($-$0.12)&
27.87 ($-$0.09)\\
\bottomrule
\end{tabular}
\vspace{-2mm}
\end{table*}

\vspace{-3mm}
\paragraph{Data augmentations.}  We trained our NAFSSR-S using different data augmentations to validate their effectiveness.  
Since we focus on data augmentation, we do not use Stochastic-Depth in this experiment.
As shown in Table~\ref{tab:aug}, the performance of NAFSSR-S is improved by introducing the data augmentation: random flip horizontally, random flip vertically, and channel shuffle mentioned in Section~\ref{sec:training}. 

When applying each data augmentation individually, the PSNR value of NAFSSR-S is improved by 0.19 dB with channel shuffle augmentation, which is compatible with random horizontal flip (+0.21 dB) and random vertical flip (+0.20 dB). This shows the effectiveness of channel shuffle augmentation. Moreover, channel shuffle is complementary to other augmentations. Using all three data augmentations boosts the PSNR value of NAFSSR-S from 23.43 dB to 23.82 dB, which is 0.09 dB better than random flip only.

\vspace{-3mm}
\paragraph{Stochastic-Depth and TLSC.}
We use NAFSSR-S and NAFSSR-B to investigate the impact of stochastic depth~\cite{huang2016deep} during training and TLSC~\cite{chu2021tlsc} during inference.
In Table~\ref{tab:droppath}, we report results on one in-distribution dataset (\ie, Flickr1024~\cite{wang2019learning} validation set) and three out-distribution datasets (\ie, 
KITTI 2012~\cite{geiger2012we}, KITTI 2015~\cite{menze2015object}, Middlebury~\cite{scharstein2014high}).

During training, stochastic depth~\cite{huang2016deep} slightly improves the performance on all datasets (+0.03 dB) for NAFSSR-S, while it improves more for larger model NAFSSR-B on both model performance (+0.11 dB on in-distribution data) and generality (+0.16 dB on out-distribution test data). 
When training without stochastic depth, NAFSSR-B performs 0.16 dB better than NAFSSR-T on Flickr1024 but only 0.07 dB better on out-distribution data. However, when using stochastic depth, NAFSSR-B outperforms NAFSSR-T on Flickr1024 and out-of-distribution data by 0.25 dB and 0.2 dB, respectively.
This shows that large models suffer from overfitting on Flickr1024 training data, while stochastic depth benefits networks and improves generality. 

During inference, TLSC~\cite{chu2021tlsc} achieves similar improvements to both NAFSSR-T and NAFSSR-B on all datasets. This indicates that NAFSSR without TLSC provides sub-optimal performance at test time due to the train-test inconsistency in stereo image SR tasks.

\subsection{Comparison to state-of-the-arts methods}

\begin{table*}[!t]
\centering
\caption{Quantitative results achieved by different methods on the KITTI 2012~\cite{geiger2012we}, KITTI 2015~\cite{menze2015object}, Middlebury~\cite{scharstein2014high}, and Flickr1024~\cite{wang2019learning} datasets. $\#P$ represents the number of parameters of the networks. 
Here, PSNR$/$SSIM values achieved on both the left images (i.e., \textit{Left}) and a pair of stereo images (i.e., $\left(\textit{Left}+\textit{Right}\right)/2$) are reported. The best results are in \textbf{bold faces}. The results of NAFSSR-L are reported only for reference (\color{gray}{gray}).
} \label{tab:sota}
\vspace{-3mm}
\resizebox{\textwidth}{!}
{
\begin{tabular}{lccccccccc}
\toprule
\multirow{2}*{Method} & \multirow{2}*{Scale} & \multirow{2}*{$\#P$} & \multicolumn{3}{c}{\textit{Left}} & \multicolumn{4}{c}{$\left(\textit{Left}+\textit{Right}\right)/2$}\\
\cmidrule(lr){4-6} \cmidrule(lr){7-10}
         &      &           & KITTI 2012 & KITTI 2015 & Middlebury & KITTI 2012 & KITTI 2015 & Middlebury & Flickr1024\\
\midrule
VDSR~\cite{kim2016accurate} & $\times$2 & 0.66M & 30.17$/$0.9062 & 28.99$/$0.9038 & 32.66$/$0.9101 & 30.30$/$0.9089 & 29.78$/$0.9150& 32.77$/$0.9102 & 25.60$/$0.8534\\
EDSR~\cite{lim2017enhanced} & $\times$2 & 38.6M & 30.83$/$0.9199 & 29.94$/$0.9231 & 34.84$/${0.9489} &30.96$/$0.9228 & 30.73$/$0.9335 & {34.95}$/${0.9492} & {28.66}$/$0.9087 \\
RDN~\cite{zhang2018residual} & $\times$2 & 22.0M  & 30.81$/$0.9197 & 29.91$/$0.9224 & {34.85}$/$0.9488 &30.94$/$0.9227 & 30.70$/$0.9330 & 34.94$/$0.9491 & 28.64$/$0.9084 \\
RCAN~\cite{zhang2018image} & $\times$2 & 15.3M & 30.88$/$0.9202 & 29.97$/$0.9231 & 34.80$/$0.9482 & 31.02$/$0.9232 & 30.77$/$0.9336 & 34.90$/$0.9486 & 28.63$/$0.9082 \\
StereoSR~\cite{jeon2018enhancing} & $\times$2 &1.08M & 29.42$/$0.9040 & 28.53$/$0.9038 & 33.15$/$0.9343 & 29.51$/$0.9073 & 29.33$/$0.9168 & 33.23$/$0.9348 & 25.96$/$0.8599 \\
PASSRnet~\cite{wang2019learning} & $\times$2 & 1.37M & 30.68$/$0.9159 & 29.81$/$0.9191 & 34.13$/$0.9421 & 30.81$/$0.9190 & 30.60$/$0.9300 & 34.23$/$0.9422 & 28.38$/$0.9038 \\
IMSSRnet~\cite{lei2020deep} & $\times$2 & 6.84M & 30.90$/$- & 29.97$/$- & 34.66$/$- & 30.92$/$- & 30.66$/$- & 34.67$/$- & -$/$- \\
iPASSR~\cite{wang2021symmetric} & $\times$2 & 1.37M & {30.97}$/${0.9210} & {30.01}$/${0.9234} & 34.41$/$0.9454 & {31.11}$/${0.9240} & {30.81}$/${0.9340} & 34.51$/$0.9454 & 28.60$/${0.9097} \\
SSRDE-FNet~\cite{dai2021feedback}   & $\times$2 & 2.10M & {31.08}$/${0.9224} & {30.10}$/${0.9245} & {35.02}$/${0.9508} & {31.23}$/${0.9254} & {30.90}$/${0.9352} & {35.09}$/${0.9511} & {28.85}$/${0.9132} \\
\midrule
NAFSSR-T (\textbf{Ours}) & $\times$2 & 0.45M  & 31.12$/$0.9224	& 30.19$/$0.9253	 & 34.93$/$0.9495 & 31.26$/$0.9254& 30.99$/$0.9355 & 35.01$/$0.9495 & 28.94$/$0.9128\\
NAFSSR-S (\textbf{Ours}) & $\times$2 & 1.54M  & 31.23$/$0.9236	& 30.28$/$0.9266	 & 35.23$/$0.9515 & 31.38$/$0.9266& 31.08$/$0.9367 & 35.30$/$0.9514 & 29.19$/$0.9160\\
NAFSSR-B (\textbf{Ours}) & $\times$2 & 6.77M  & \textbf{31.40}$/$\textbf{0.9254} & \textbf{30.42}$/$\textbf{0.9282} & \textbf{35.62}$/$\textbf{0.9545} & \textbf{31.55}$/$\textbf{0.9283} & \textbf{31.22}$/$\textbf{0.9380} & \textbf{35.68}$/$\textbf{0.9544} & \textbf{29.54}$/$\textbf{0.9204} \\
\textcolor{gray}{NAFSSR-L (\textbf{Ours})} & \textcolor{gray}{$\times$2} & \textcolor{gray}{23.79M}  & \textcolor{gray}{\textbf{31.45}}$/$\textcolor{gray}{\textbf{0.9261}} & \textcolor{gray}{\textbf{30.46}}$/$\textcolor{gray}{\textbf{0.9289}} & \textcolor{gray}{\textbf{35.83}}$/$\textcolor{gray}{\textbf{0.9559}} & \textcolor{gray}{\textbf{31.60}}$/$\textcolor{gray}{\textbf{0.9291}} & \textcolor{gray}{\textbf{31.25}}$/$\textcolor{gray}{\textbf{0.9386}} & \textcolor{gray}{\textbf{35.88}}$/$\textcolor{gray}{\textbf{0.9557}} & \textcolor{gray}{\textbf{29.68}}$/$\textcolor{gray}{\textbf{0.9221}} \\
\midrule
\midrule
VDSR~\cite{kim2016accurate} &  $\times$4 & 0.66M & 25.54$/$0.7662 & 24.68$/$0.7456 & 27.60$/$0.7933 & 25.60$/$0.7722 & 25.32$/$0.7703 & 27.69$/$0.7941 & 22.46$/$0.6718 \\
EDSR~\cite{lim2017enhanced} &  $\times$4 & 38.9M & 26.26$/$0.7954 & 25.38$/$0.7811 & 29.15$/${0.8383} & 26.35$/$0.8015 & 26.04$/$0.8039 & 29.23$/$0.8397 & 23.46$/$0.7285 \\
RDN~\cite{zhang2018residual} &  $\times$4 & 22.0M  & 26.23$/$0.7952 & 25.37$/$0.7813 & 29.15$/$0.8387 & 26.32$/$0.8014 & 26.04$/$0.8043 & 29.27$/${0.8404} & 23.47$/${0.7295} \\
RCAN~\cite{zhang2018image} &  $\times$4 & 15.4M & 26.36$/$0.7968 & 25.53$/$0.7836 & {29.20}$/$0.8381 & 26.44$/$0.8029 & 26.22$/$0.8068 & {29.30}$/$0.8397 & {23.48}$/$0.7286 \\
StereoSR~\cite{jeon2018enhancing}  &  $\times$4 & 1.42M   & 24.49$/$0.7502 & 23.67$/$0.7273 &27.70$/$0.8036 & 24.53$/$0.7555 & 24.21$/$0.7511 & 27.64$/$0.8022 & 21.70$/$0.6460 \\
PASSRnet~\cite{wang2019learning}  &  $\times$4 & 1.42M   & 26.26$/$0.7919 & 25.41$/$0.7772 &28.61$/$0.8232 & 26.34$/$0.7981 & 26.08$/$0.8002 & 28.72$/$0.8236 & 23.31$/$0.7195 \\
SRRes+SAM~\cite{ying2020stereo}  &  $\times$4 & 1.73M  & 26.35$/$0.7957 & 25.55$/$0.7825 & 28.76$/$0.8287 & 26.44$/$0.8018 & 26.22$/$0.8054 & 28.83$/$0.8290 & 23.27$/$0.7233 \\
IMSSRnet~\cite{lei2020deep} &  $\times$4 & 6.89M  & 26.44$/$- & 25.59$/$- & 29.02$/$- & 26.43$/$- & 26.20$/$- & 29.02$/$- & -$/$- \\
iPASSR~\cite{wang2021symmetric}  &  $\times$4 & 1.42M  & {26.47}$/${0.7993} & {25.61}$/${0.7850} & 29.07$/$0.8363 & {26.56}$/${0.8053} & {26.32}$/${0.8084} & 29.16$/$0.8367 & 23.44$/$0.7287 \\
SSRDE-FNet~\cite{dai2021feedback}  & $\times$4 & 2.24M  & {26.61}$/${0.8028} & {25.74}$/${0.7884} & {29.29}$/${0.8407} & {26.70}$/${0.8082} & {26.43}$/${0.8118} & {29.38}$/${0.8411} & {23.59}$/${0.7352} \\
\midrule
NAFSSR-T (\textbf{Ours}) & $\times$4 & 0.46M  & 26.69$/$0.8045	& 25.90$/$0.7930	 & 29.22$/$0.8403 & 26.79$/$0.8105& 26.62$/$0.8159 & 29.32$/$0.8409 & 23.69$/$0.7384\\
NAFSSR-S (\textbf{Ours}) & $\times$4 & 1.56M  & 26.84$/$0.8086	& 26.03$/$0.7978	 & 29.62$/$0.8482 & 26.93$/$0.8145& 26.76$/$0.8203 & 29.72$/$0.8490 & 23.88$/$0.7468\\
NAFSSR-B (\textbf{Ours}) & $\times$4 & 6.80M  & \textbf{26.99}$/$\textbf{0.8121} & \textbf{26.17}$/$\textbf{0.8020} & \textbf{29.94}$/$\textbf{0.8561} & \textbf{27.08}$/$\textbf{0.8181} & \textbf{26.91}$/$\textbf{0.8245} & \textbf{30.04}$/$\textbf{0.8568} & \textbf{24.07}$/$\textbf{0.7551} \\
\textcolor{gray}{NAFSSR-L (\textbf{Ours})} & \textcolor{gray}{$\times$4} & \textcolor{gray}{23.83M}  & \textcolor{gray}{\textbf{27.04}}$/$\textcolor{gray}{\textbf{0.8135}} & \textcolor{gray}{\textbf{26.22}}$/$\textcolor{gray}{\textbf{0.8034}} & \textcolor{gray}{\textbf{30.11}}$/$\textcolor{gray}{\textbf{0.8601}} & \textcolor{gray}{\textbf{27.12}}$/$\textcolor{gray}{\textbf{0.8194}} & \textcolor{gray}{\textbf{26.96}}$/$\textcolor{gray}{\textbf{0.8257}} & \textcolor{gray}{\textbf{30.20}}$/$\textcolor{gray}{\textbf{0.8605}} & \textcolor{gray}{\textbf{24.17}}$/$\textcolor{gray}{\textbf{0.7589}} \\
\bottomrule
\end{tabular}}
\vspace{-3mm}
\end{table*}

\input{figures/Flickr1024}

\begin{table}[]
\centering
\caption{PSNR v.s. runtimes on Flickr1024 dataset for $4\times$ SR.
}
\vspace{-3mm}
\label{tab:runtime}
\small
\begin{tabular}{cccc}
\toprule
Models    & PSNR & Time(ms) & Speedup \\ \midrule
SSRDEFNet~\cite{dai2021feedback}  & 23.59      & 238.5    & 1.00$\times$       \\ \midrule
NAFSSR-T (Ours)  & 23.64 (+0.05)     & 46.7     & 5.11$\times$   \\
NAFSSR-S (Ours)  & 23.88 (+0.29)        & 91.8     & 2.60$\times$    \\
NAFSSR-B (Ours)  & 24.07 (+0.48)     & 224.9    & 1.06$\times$   \\
\bottomrule
\end{tabular}
\vspace{-2mm}
\end{table}

\subsubsection{Settings} 
\textbf{Training data.} We use training data that are identical to iPASSR~\cite{wang2021symmetric} to provide a fair comparison with previous work. In detail, the 800 images from training set of Flickr1024~\cite{wang2019flickr1024} and 60 Middlebury~\cite{scharstein2014high} images are used for training.
Following~\cite{wang2021symmetric}, we perform bicubic downsampling by a factor of 2 on images from the Middlebury dataset to generate high-resolution (HR) ground truth images so that they match the spatial resolution of the Flickr1024 dataset.
To produce low-resolution images, we apply bicubic downsampling to HR images on specific scaling factors (\ie, $2\times$ and $4\times$) and then crop $30\times90$ patches with a stride of 20 as inputs. Limited by the size of the offline cropped patches, we do not use additional random crop in this section. 

\textbf{Evaluation details.} 
To evaluate SR results, 20 images from KITTI 2012~\cite{geiger2012we} and 20 images from KITTI 2015~\cite{menze2015object}, 5 images from Middlebury~\cite{scharstein2014high}, and 112 images from the test set of Flickr1024~\cite{wang2019learning} are utilized for testing. 
Note that different from Section~\ref{sec:NTIRE.details}, the test images used in this section are from the test set instead of the validation set of Flickr1024 dataset.
Following~\cite{wang2021symmetric}, we report PSNR/SSIM scores on the left images with their left boundaries (64 pixels) cropped, and average scores on stereo image pairs (\ie, (Left + Right) /2) without any boundary cropping.
\input{figures/KITTI}

\input{figures/Middlebury}

\vspace{-1mm}
\subsubsection{Results}
We compare our NAFSSR (with 3 different variants) with existing super-resolution (SR) methods, including single image SR methods (\ie, VDSR~\cite{kim2016accurate}, EDSR~\cite{lim2017enhanced}, RDN~\cite{zhang2018residual}, and RCAN~\cite{zhang2018image}) and stereo image SR methods (\ie, StereoSR~\cite{jeon2018enhancing}, PASSRnet~\cite{wang2019learning}, SRRes+SAM~\cite{ying2020stereo}, IMSSRnet~\cite{lei2020deep}, iPASSR~\cite{wang2021symmetric} and SSRDE-FNet~\cite{dai2021feedback}). This methods are trained on the same training datasets as ours and their PSNR and SSIM scores are reported by~\cite{dai2021feedback}.

\textbf{Quantitative Evaluations.}
The quantitative comparisons with existing SR methods are shown in Table~\ref{tab:sota}. Our smallest NAFSSR-T achieves competitive results as previous state-of-the-art (SSRDE-FNet~\cite{dai2021feedback}), and our NAFSSR-S outperforms the state-of-the-art results on all datasets and upsampling factors ($\times$2, $\times$4). Furthermore, our NAFSSR-B improves state-of-the-art results of all datasets by a significant margin.
For example, for $4\times$ stereo SR, our NAFSSR-B surpass previous state-of-the-art model SSRDE-FNet~\cite{dai2021feedback} by 0.38 dB, 0.48 dB, 0.66 dB, 0.48 dB on KITTI 2012~\cite{geiger2012we}, KITTI 2015~\cite{menze2015object}, Middlebury~\cite{scharstein2014high} and Flickr1024~\cite{wang2019learning}, respectively. This clearly shows the effectiveness of the proposed NAFSSR.

\textbf{Parameter Efficiency and Scaling Ability.} We also visualize the trade-off results between total numbers of parameters and PSNR on Flickr1024 dataset for 4$\times$ stereo SR. As shown in Figure~\ref{fig:nafssr-params}, compared with SSRDE-FNet~\cite{dai2021feedback}, our NAFSSR-T achieves state-of-the-art result with 79\% parameter reduction. This shows that NAFSSR has high parameter efficiency. 
Furthermore, by scaling up the model size, our NAFSSR-S clearly surpasses competitive methods with similar total numbers of parameters, and NAFSSR-B and NAFSSR-L further push the state-of-the-art stereo SR performance. This shows the scaling ability of NAFSSR.

\textbf{Runtime Efficiency.} 
We also report the runtimes (evaluated with $128\times128$ input on RTX 2080Ti GPU) to compare the computational complexity between existing best model SSRDE-FNet~\cite{dai2021feedback} and our NAFSSR. As shown in Table~\ref{tab:runtime}, all variants of NAFSSR outperform SSRDE-FNet by a PSNR margin of 0.05 $\sim$ 0.48 dB
on Flickr1024~\cite{wang2019learning} dataset, with up to 5.11$\times$ speedup.
This indicates that the NAFSSR architecture is fast and efficient.

\textbf{Visual Comparison.}
In Figures~\ref{fig:flickr1024}, \ref{fig:kitti} and \ref{fig:middlebury}, we show the visual comparisons for $\times 4$ stereo SR on Flickr1024~\cite{wang2019learning}, KITTI 2012~\cite{geiger2012we}, KITTI 2015~\cite{menze2015object} and  Middlebury~\cite{scharstein2014high}. 
These figures show that our NAFSSR-B reconstructs pleasing SR images with rich details and clear edges. In contrast, other compared methods may suffer from unsatisfactory artifacts.
This confirms the effectiveness of our NAFSSR.

%% file: figures/Flickr1024.tex
\begin{figure*}[t]
\scriptsize
\centering
\begin{tabular}{cc}
\hspace{-0.4cm}
\begin{adjustbox}{valign=t}
\begin{tabular}{c}
\includegraphics[width=0.152\textwidth]{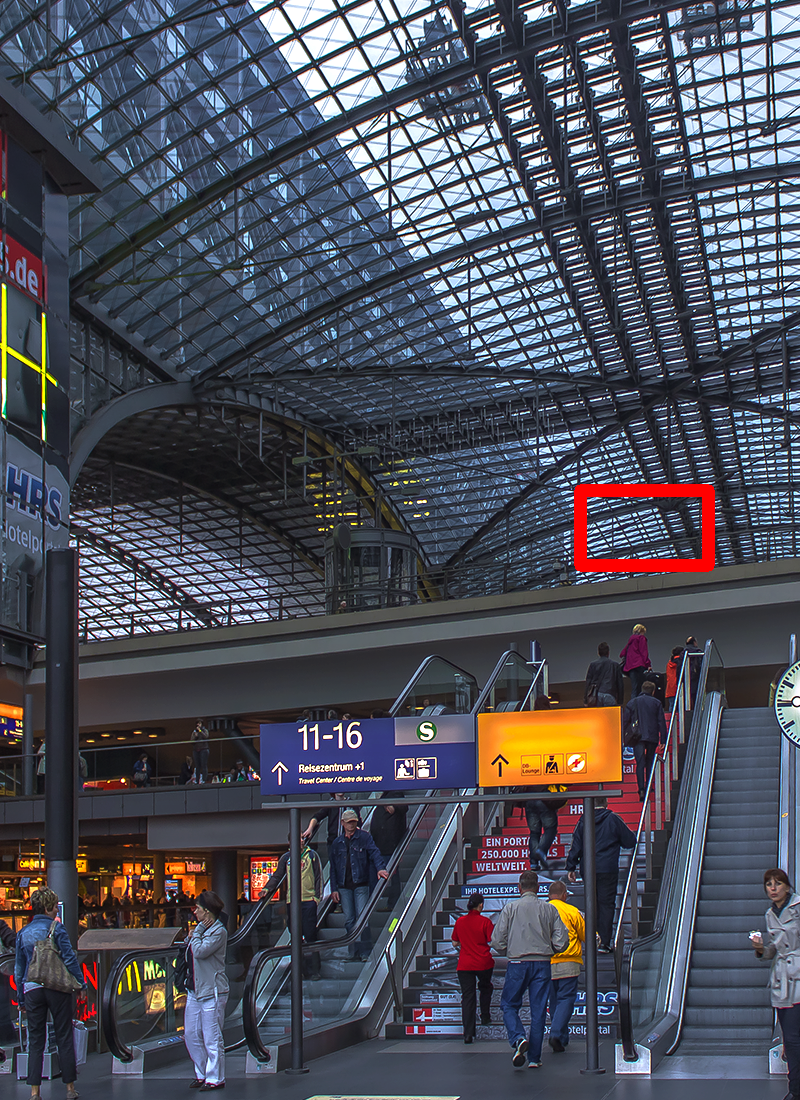}
\\
img\_0035 (Left)
\end{tabular}
\end{adjustbox}
\hspace{-0.46cm}
\begin{adjustbox}{valign=t}
\begin{tabular}{cccccc}
\includegraphics[width=0.161\textwidth]{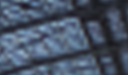} \hspace{-4mm} &
\includegraphics[width=0.161\textwidth]{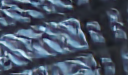} \hspace{-4mm} &
\includegraphics[width=0.161\textwidth]{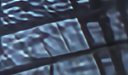} \hspace{-4mm} &
\includegraphics[width=0.161\textwidth]{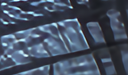} \hspace{-4mm} &
\includegraphics[width=0.161\textwidth]{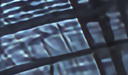} \hspace{-4mm} 
\\

Bicubic \hspace{-4mm} &
StereoSR~\cite{jeon2018enhancing} \hspace{-4mm} &
EDSR~\cite{lim2017enhanced} \hspace{-4mm} &
RDN~\cite{zhang2018residual} \hspace{-4mm} &
RCAN~\cite{zhang2018image} \hspace{-4mm}
\\

\includegraphics[width=0.161\textwidth]{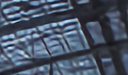} \hspace{-4mm} &
\includegraphics[width=0.161\textwidth]{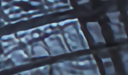} \hspace{-4mm} &
\includegraphics[width=0.161\textwidth]{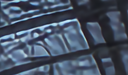} \hspace{-4mm} &
\includegraphics[width=0.161\textwidth]{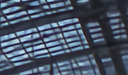} \hspace{-4mm}   &
\includegraphics[width=0.161\textwidth]{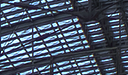} \hspace{-4mm} 
\\ 
SRRes+SAM~\cite{ying2020stereo} \hspace{-4mm} &
iPASSR~\cite{wang2021symmetric} \hspace{-4mm} &
SSRDE-FNet~\cite{dai2021feedback}  \hspace{-4mm} &
NAFSSR-B (ours) \hspace{-4mm} &
Reference \hspace{-4mm}
\\
\end{tabular}
\end{adjustbox}
\vspace{1mm}
\\

\hspace{-0.4cm}
\begin{adjustbox}{valign=t}
\begin{tabular}{c}
\includegraphics[width=0.152\textwidth]{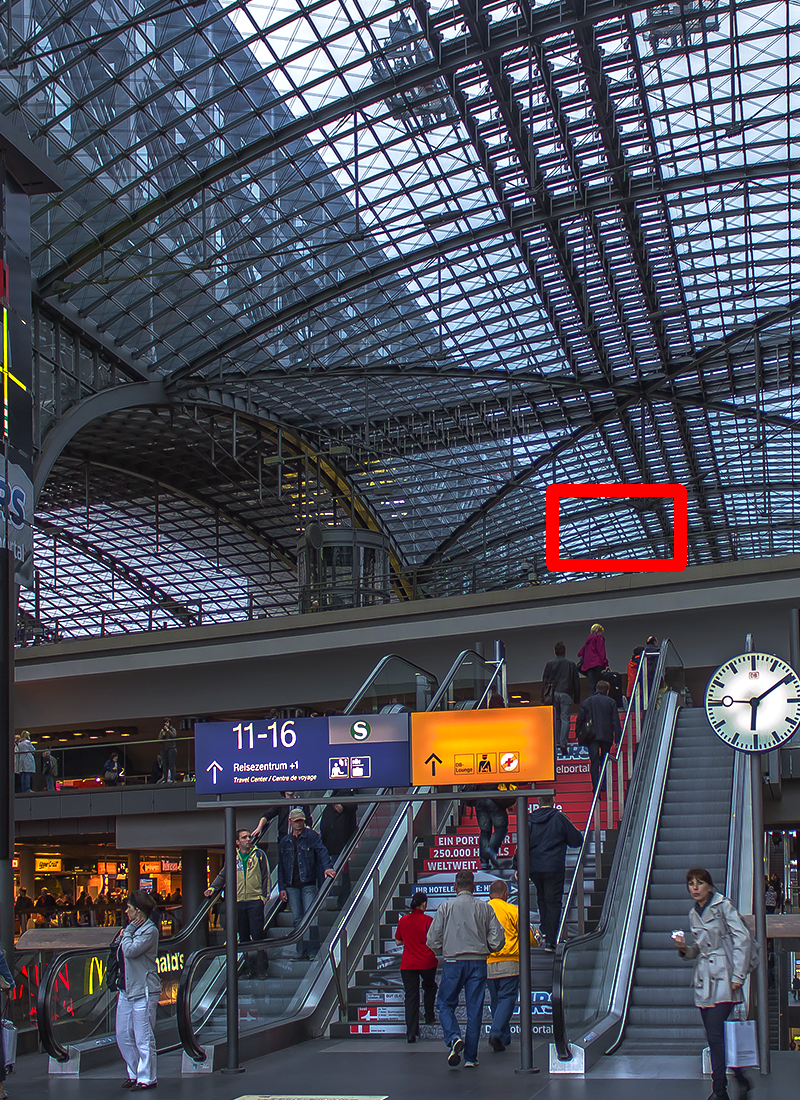}
\\
img\_0035 (Right)
\end{tabular}
\end{adjustbox}
\hspace{-0.46cm}
\begin{adjustbox}{valign=t}
\begin{tabular}{cccccc}

\includegraphics[width=0.161\textwidth]{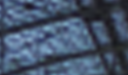} \hspace{-4mm} &
\includegraphics[width=0.161\textwidth]{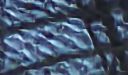} \hspace{-4mm} &
\includegraphics[width=0.161\textwidth]{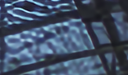} \hspace{-4mm} &
\includegraphics[width=0.161\textwidth]{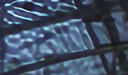} \hspace{-4mm} &
\includegraphics[width=0.161\textwidth]{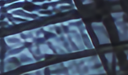} \hspace{-4mm} 
\\

Bicubic \hspace{-4mm} &
StereoSR~\cite{jeon2018enhancing} \hspace{-4mm} &
EDSR~\cite{lim2017enhanced} \hspace{-4mm} &
RDN~\cite{zhang2018residual} \hspace{-4mm} &
RCAN~\cite{zhang2018image} \hspace{-4mm}
\\

\includegraphics[width=0.161\textwidth]{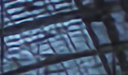} \hspace{-4mm} &
\includegraphics[width=0.161\textwidth]{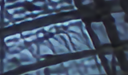} \hspace{-4mm} &
\includegraphics[width=0.161\textwidth]{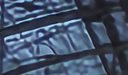} \hspace{-4mm} &
\includegraphics[width=0.161\textwidth]{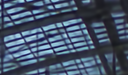} \hspace{-4mm}   &
\includegraphics[width=0.161\textwidth]{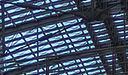} \hspace{-4mm} 
\\ 

SRRes+SAM~\cite{ying2020stereo} \hspace{-4mm} &
iPASSR~\cite{wang2021symmetric} \hspace{-4mm} &
SSRDE-FNet~\cite{dai2021feedback}  \hspace{-4mm} &
NAFSSR-B (ours) \hspace{-4mm} &
Reference \hspace{-4mm} 
\\
\end{tabular}
\end{adjustbox}
\vspace{1mm}
\\
\end{tabular}
\vspace{-3mm}
\caption{Visual results ($\times$4) achieved by different methods on the  Flickr1024~\cite{wang2019learning} dataset. 
}
\label{fig:flickr1024}
\vspace{-3mm}
\end{figure*}

%% file: figures/KITTI.tex
\begin{figure*}[!t]
\begin{center}
\scalebox{0.99}{
\begin{tabular}[b]{c@{ } c@{ } c@{ }  c@{ } c@{ } c@{ } c@{ } }
\includegraphics[width=.136\textwidth,valign=t]{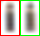} &
    \includegraphics[width=.136\textwidth,valign=t]{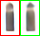} &
    \includegraphics[width=.136\textwidth,valign=t]{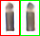} &
    \includegraphics[width=.136\textwidth,valign=t]{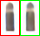} &
    \includegraphics[width=.136\textwidth,valign=t]{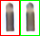} &
    \includegraphics[width=.136\textwidth,valign=t]{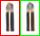} &
    \includegraphics[width=.136\textwidth,valign=t]{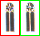}
\\
\scriptsize~Bicubic    &
\scriptsize~RCAN~\cite{zhang2018image} &
\scriptsize~SRRes+SAM~\cite{ying2020stereo} &\scriptsize~iPASSR~\cite{wang2021symmetric} &\scriptsize~SSRDE-FNet~\cite{dai2021feedback} &\scriptsize~NAFSSR-B (ours)  & \scriptsize~Reference\\
 \includegraphics[width=.136\textwidth,valign=t]{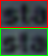} &
 \includegraphics[width=.136\textwidth,valign=t]{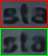}&
    \includegraphics[width=.136\textwidth,valign=t]{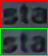} &
    
    \includegraphics[width=.136\textwidth,valign=t]{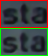} &
    \includegraphics[width=.136\textwidth,valign=t]{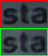} &
    \includegraphics[width=.136\textwidth,valign=t]{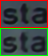} &
    \includegraphics[width=.136\textwidth,valign=t]{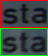} \\
    \scriptsize~Bicubic    &
\scriptsize~RCAN~\cite{zhang2018image} &
\scriptsize~SRRes+SAM~\cite{ying2020stereo} &\scriptsize~iPASSR~\cite{wang2021symmetric} &\scriptsize~SSRDE-FNet~\cite{dai2021feedback} &\scriptsize~NAFSSR-B (ours)  & \scriptsize~Reference\\
\end{tabular}}
\end{center}
\vspace{-7mm}
\caption{Visual results ($\times$4) achieved by different methods on the KITTI 2012~\cite{geiger2012we} (top) and KITTI 2015~\cite{menze2015object} (bottom) dataset. The images with red and green borders represent the left and right views respectively.
}
\label{fig:kitti}
\vspace{-3mm}
\end{figure*}

%% file: figures/Middlebury.tex
\begin{figure*}[t]
\scriptsize
\centering
\begin{tabular}{cc}
\hspace{-0.4cm}
\begin{adjustbox}{valign=t}
\begin{tabular}{c}
\includegraphics[width=0.178\textwidth]{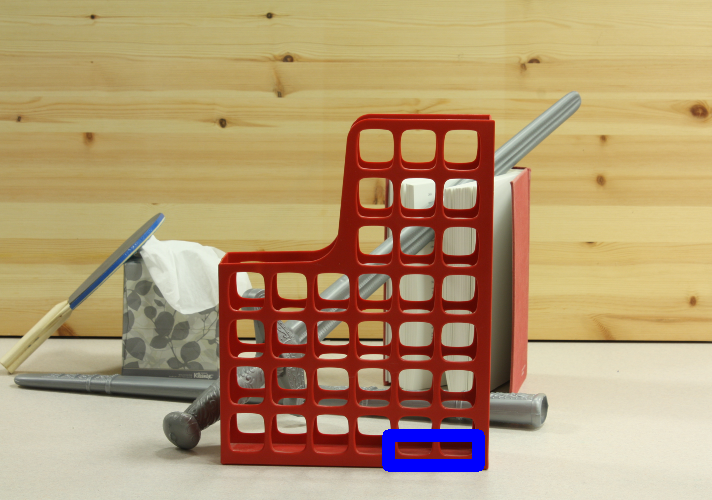}
\\
img\_sword2 (Left)
\end{tabular}
\end{adjustbox}
\hspace{-0.46cm}
\begin{adjustbox}{valign=t}
\begin{tabular}{cccccc}

\includegraphics[width=0.156\textwidth]{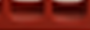} \hspace{-4mm} &
\includegraphics[width=0.156\textwidth]{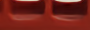} \hspace{-4mm} &
\includegraphics[width=0.156\textwidth]{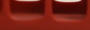} \hspace{-4mm} &
\includegraphics[width=0.156\textwidth]{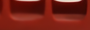} \hspace{-4mm} &
\includegraphics[width=0.156\textwidth]{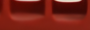} \hspace{-4mm} 
\\
Bicubic \hspace{-4mm} &
StereoSR~\cite{jeon2018enhancing} \hspace{-4mm} &
EDSR~\cite{lim2017enhanced} \hspace{-4mm} &
RDN~\cite{zhang2018residual} \hspace{-4mm} &
RCAN~\cite{zhang2018image} \hspace{-4mm}
\\

\includegraphics[width=0.156\textwidth]{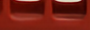} \hspace{-4mm} &
\includegraphics[width=0.156\textwidth]{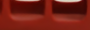} \hspace{-4mm} &
\includegraphics[width=0.156\textwidth]{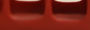} \hspace{-4mm} &
\includegraphics[width=0.156\textwidth]{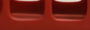} \hspace{-4mm}   &
\includegraphics[width=0.156\textwidth]{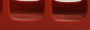} \hspace{-4mm} 
\\ 
SRRes+SAM~\cite{ying2020stereo} \hspace{-4mm} &
iPASSR~\cite{wang2021symmetric} \hspace{-4mm} &
SSRDE-FNet~\cite{dai2021feedback}  \hspace{-4mm} &
NAFSSR-B (ours) \hspace{-4mm} &
Reference \hspace{-4mm} 
\\
\end{tabular}
\end{adjustbox}
\vspace{1mm}
\\

\hspace{-0.4cm}
\begin{adjustbox}{valign=t}
\begin{tabular}{c}
\includegraphics[width=0.178\textwidth]{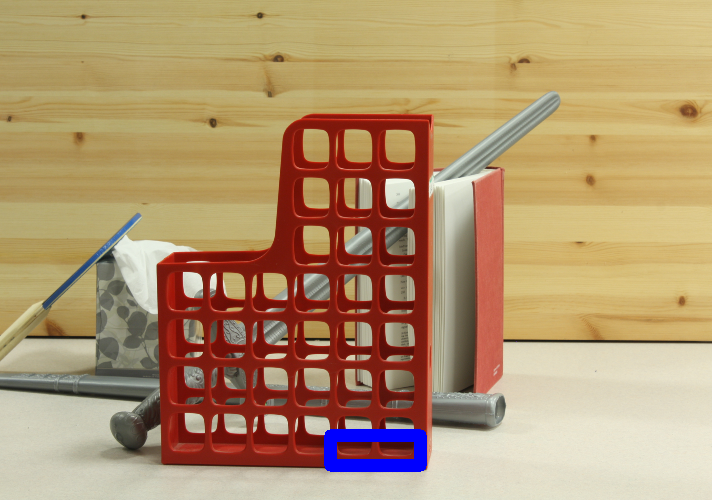}
\\
img\_sword2 (Right)
\end{tabular}
\end{adjustbox}
\hspace{-0.46cm}
\begin{adjustbox}{valign=t}
\begin{tabular}{cccccc}
\includegraphics[width=0.156\textwidth]{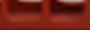} \hspace{-4mm} &
\includegraphics[width=0.156\textwidth]{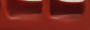} \hspace{-4mm} &
\includegraphics[width=0.156\textwidth]{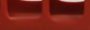} \hspace{-4mm} &
\includegraphics[width=0.156\textwidth]{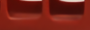} \hspace{-4mm} &
\includegraphics[width=0.156\textwidth]{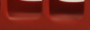} \hspace{-4mm} 
\\
Bicubic \hspace{-4mm} &
StereoSR~\cite{jeon2018enhancing} \hspace{-4mm} &
EDSR~\cite{lim2017enhanced} \hspace{-4mm} &
RDN~\cite{zhang2018residual} \hspace{-4mm} &
RCAN~\cite{zhang2018image} \hspace{-4mm}
\\

\includegraphics[width=0.156\textwidth]{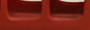} \hspace{-4mm} &
\includegraphics[width=0.156\textwidth]{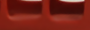} \hspace{-4mm} &
\includegraphics[width=0.156\textwidth]{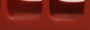} \hspace{-4mm} &
\includegraphics[width=0.156\textwidth]{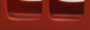} \hspace{-4mm}   &
\includegraphics[width=0.156\textwidth]{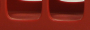} \hspace{-4mm} 
\\  

SRRes+SAM~\cite{ying2020stereo} \hspace{-4mm} &
iPASSR~\cite{wang2021symmetric} \hspace{-4mm} &
SSRDE-FNet~\cite{dai2021feedback}  \hspace{-4mm} &
NAFSSR-B (ours) \hspace{-4mm} &
Reference \hspace{-4mm} 
\\
\end{tabular}
\end{adjustbox}
\vspace{1mm}
\\

\vspace{-6mm}
\end{tabular}
\caption{Visual results ($\times$4) achieved by different methods on the  Middlebury~\cite{scharstein2014high} dataset. 
}
\label{fig:middlebury}
\vspace{-3mm}
\end{figure*}

%% file: chaps/challenge.tex
\subsection{NTIRE Stereo Image SR Challenge}
We submitted a result obtained by the presented approach
to the NTIRE 2022 Stereo Image Super-Resolution Challenge~\cite{Wang2022NTIRE}.
In order to maximize the potential performance of our method, we further enlarge the NAFSSR-Base by increasing its depth and width. We adopt stronger stochastic depth~\cite{huang2016deep} with $0.3$ or $0.4$ probability to overcome the overfitting issue. During test-time, we adopt both self-ensemble~\cite{lim2017enhanced} and model ensemble strategy. Specifically, the data augmentations mentioned in Section~\ref{sec:training} are used as test-time data augmentations for self-ensemble. Inspired by~\cite{wortsman2022model}, we further ensemble multiple models trained with various hyper-parameters. As a result, our final submission achieves 24.239 dB PSNR on the validation set and won the first place with 23.787 dB PSNR on the test set.

%% file: chaps/conclusion.tex
\section{Conclusion}
This paper proposes a simple baseline named NAFSSR for stereo image super-resolution (SR). 
We use a stack of NAFBlock for intra-view feature extraction and combine it with stereo cross attention modules for cross-view feature interaction. Furthermore, we adopt stronger data augmentations for training and solve the train-test inconsistency in stereo image SR tasks by the test-time local converter.
We also employ stochastic depth technique to improve the generality of large models. 
Extensive experiments show that NAFSSR surpasses current models and achieves state-of-the-art performance.